\documentclass{article}

\usepackage{microtype}
\usepackage{graphicx}
\usepackage{booktabs}
\usepackage{hyperref}
\usepackage{xurl}

\usepackage[preprint]{icml2027}

\usepackage{amsmath,amssymb,bm}
\usepackage{array}

\icmltitlerunning{LACE: Compact Experts for Irregular Time-Series Forecasting}

\begin{document}

\twocolumn[
  \icmltitle{LACE: LLM-Anchored Compact Experts for Irregular Time-Series Forecasting}
  \begin{icmlauthorlist}
    \icmlauthor{Shusen Ma}{anon}
    \icmlauthor{Xudong Zuo}{anon}
    \icmlauthor{Yangchun Xie}{anon}
    \icmlauthor{Yun-Bo Zhao}{anon}
  \end{icmlauthorlist}
  \icmlaffiliation{anon}{University of Science and Technology of China}
  \icmlcorrespondingauthor{Yun-Bo Zhao}{ybzhao@ustc.edu.cn}
  \icmlkeywords{Irregular time series forecasting, Large language models, Mixture of experts}
  \vskip 0.3in
]
\begin{NoHyper}\printAffiliationsAndNotice{}\end{NoHyper}

\begin{abstract}
Measurements in a multivariate time series often arrive on different schedules. A forecaster must decide how much to learn from each channel's history and when observations of another channel can help. We propose LACE (LLM-Anchored Compact Experts), which shares a small set of temporal forecasting bases across channels. A frozen large language model (LLM) encodes training-history summaries and available channel metadata into low-rank coordinates. The forecaster uses these coordinates to adapt expert coefficients and to exchange standardized forecast changes across channels. Both operations are gated by the observed history. The LLM runs during offline compilation. The resulting forecaster has 1,992--6,867 trainable parameters. Compared with thirteen baselines, LACE has the lowest mean squared error (MSE) on all four benchmarks and the lowest mean absolute error (MAE) on three, averaged over five seeds. It ranks second on USHCN MAE. Relative to APN, mean MSE falls by 4.01\% on HumanActivity, 2.64\% on USHCN, 2.87\% on PhysioNet, and 5.33\% on MIMIC-III.
\end{abstract}
\section{Introduction}
\label{sec:lace-introduction}

In an intensive-care record, one variable may have several recent measurements while another was last measured hours ago. Both observation patterns matter when forecasting the patient's next measurements. Closely spaced readings can reveal a local trend, whereas a long gap makes that trend harder to extrapolate. A recent change in a related variable may then be more informative than the target variable's own last two readings.

Forecasting models for irregular multivariate series use timestamps and channel identities in different ways. GraFITi represents a measurement as an edge between a time node and a channel node~\cite{yalavarthi2024grafiti}. APN learns temporal patches that summarize observations at different scales~\cite{liu2026apn}. We ask how a compact forecaster can share a set of temporal patterns across variables while allowing each variable to draw on different patterns and different neighbors.

Large language models (LLMs) offer one source of information about how variables relate. Time-LLM maps numerical patches to text prototypes, and $S^2$IP-LLM retrieves semantic anchors for its forecasting prompts~\cite{jin2024timellm,pan2024s2ip}. Both methods run the LLM at prediction time. In several such systems, replacing the LLM produced comparable or lower forecast error~\cite{tan2024useful}. This finding motivates a more specific role for the LLM. It can describe the channels once, leaving a numerical model to decide which descriptions are useful for a particular forecast.

We propose \textbf{LACE (LLM-Anchored Compact Experts)} for irregular time-series forecasting. A frozen LLM reads summaries of each channel's training histories together with available metadata. We extract a low-rank coordinate vector for each channel and store it with the forecaster. These coordinates guide two decisions. First, they adapt the coefficients of a shared bank of temporal bases to the target channel. A gate adjusts the size of the adaptation using the amount and timing of the observed data. Second, the coordinates guide the exchange of predicted changes between channels. Each source expresses a change relative to its last observation and in units of its historical variation. LACE combines these changes using channel similarity and observation reliability, then rescales the result to the target channel. A numerical router mixes the experts for each query, and a low-rank correction uses the last observed values to adjust the forecast level.

\begin{figure}[h]
\centering
\includegraphics[width=\columnwidth]{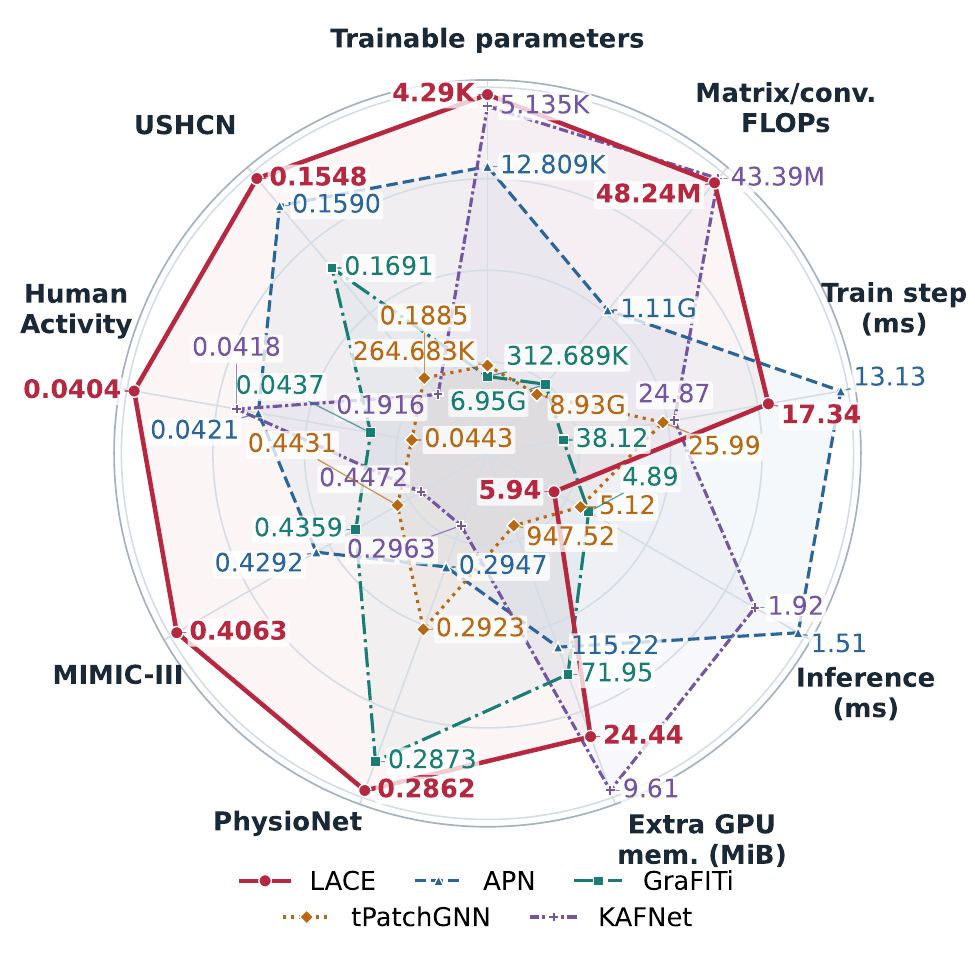}
\caption{\textbf{Forecast accuracy and computational cost.}
The four dataset axes show Table~\ref{tab:accuracy} MSE. Each cost is the mean across datasets for batches of 32 on an RTX 4090 D. K, M, and G denote $10^3$, $10^6$, and $10^9$. Each axis is log-scaled, with lower values farther out. Appendix~\ref{app:lace-efficiency} gives the measurement protocol and per-dataset costs.}
\label{fig:lace-efficiency}
\end{figure}

Figure~\ref{fig:lace-efficiency} compares these errors with the cost of forecasting. LACE averages 4.29K trainable parameters across the four datasets, about one-third of APN's 12.81K. APN has the shortest inference time among the five models shown, averaging 1.51\,ms per batch to LACE's 5.94\,ms.

Our contributions are threefold.
\begin{itemize}
 \item We propose LACE, a compact expert model for irregular forecasting. The experts combine fixed temporal bases at each query time, with 1,992--6,867 trainable parameters across the four benchmarks.
 \item We introduce an offline LLM representation for two forms of channel sharing. Each channel's coordinates adapt its expert coefficients and weight forecast changes from other observed channels. The observed history gates both contributions.
 \item Extensive experiments show that LACE achieves the lowest mean MSE on all four benchmarks and the lowest mean MAE on three.
\end{itemize}
\section{Related Work}
\label{sec:lace-related-work}

\paragraph{Irregular observations and channel interactions.}
Latent ODEs and continuous recurrent units evolve a hidden state between observations~\citep{rubanova2019latentode,schirmer2022cru}. Neural CDEs drive the hidden state with an interpolated input path, and ContiFormer extends attention to continuous time~\citep{kidger2020cde,chen2023contiformer}. Multi-Time Attention Networks aggregate observations at reference times, and DNBNet learns temporal bases and their importance weights~\citep{shukla2021mtan,li2026dnbnet}. GraFITi builds a bipartite graph of channels and observation times, tPatchGNN passes messages across aligned patches, and APN learns temporal patch boundaries~\citep{yalavarthi2024grafiti,zhang2024tpatchgnn,liu2026apn}. KAFNet applies temporal kernels to pre-aligned series and models cross-channel dependence in the frequency domain~\citep{zhou2026kafnet}. LACE summarizes each channel with fixed event-time bases and exchanges standardized changes across channels at each forecast query.

\paragraph{Language models for forecasting.}
Time-LLM maps numerical patches to text prototypes, and $S^2$IP-LLM retrieves semantic anchors for forecasting prompts~\citep{jin2024timellm,pan2024s2ip}. TEMPO uses prompts with trend, seasonal, and residual components~\citep{cao2024tempo}. Chronos trains language-model architectures on quantized time series~\citep{ansari2024chronos}, and TimeKD distills future-aware teacher representations into a smaller forecaster~\citep{liu2025timekd}. Ablations of several LLM-based forecasters have found comparable accuracy after replacing or removing the LLM~\citep{tan2024useful}. In LACE, the teacher produces a fixed representation of each channel. Retraining the forecaster with these representations removed or reassigned to other channels measures how much the forecasts depend on them.

\paragraph{Forecasting experts.}
MoLE learns a mixture of linear experts, while LeMoLE uses language-model features to modulate experts with different lookback windows~\citep{ni2024mole,zhang2024lemole}. Time-MoE routes inputs to a sparse set of experts in a pretrained autoregressive forecaster~\citep{shi2025timemoe}. LACE uses channel representations to adapt the coefficients of experts that share a temporal basis. The numerical history and query time determine their mixture weights.
\begin{figure*}[!t]
  \centering
  \includegraphics[width=.87\textwidth]{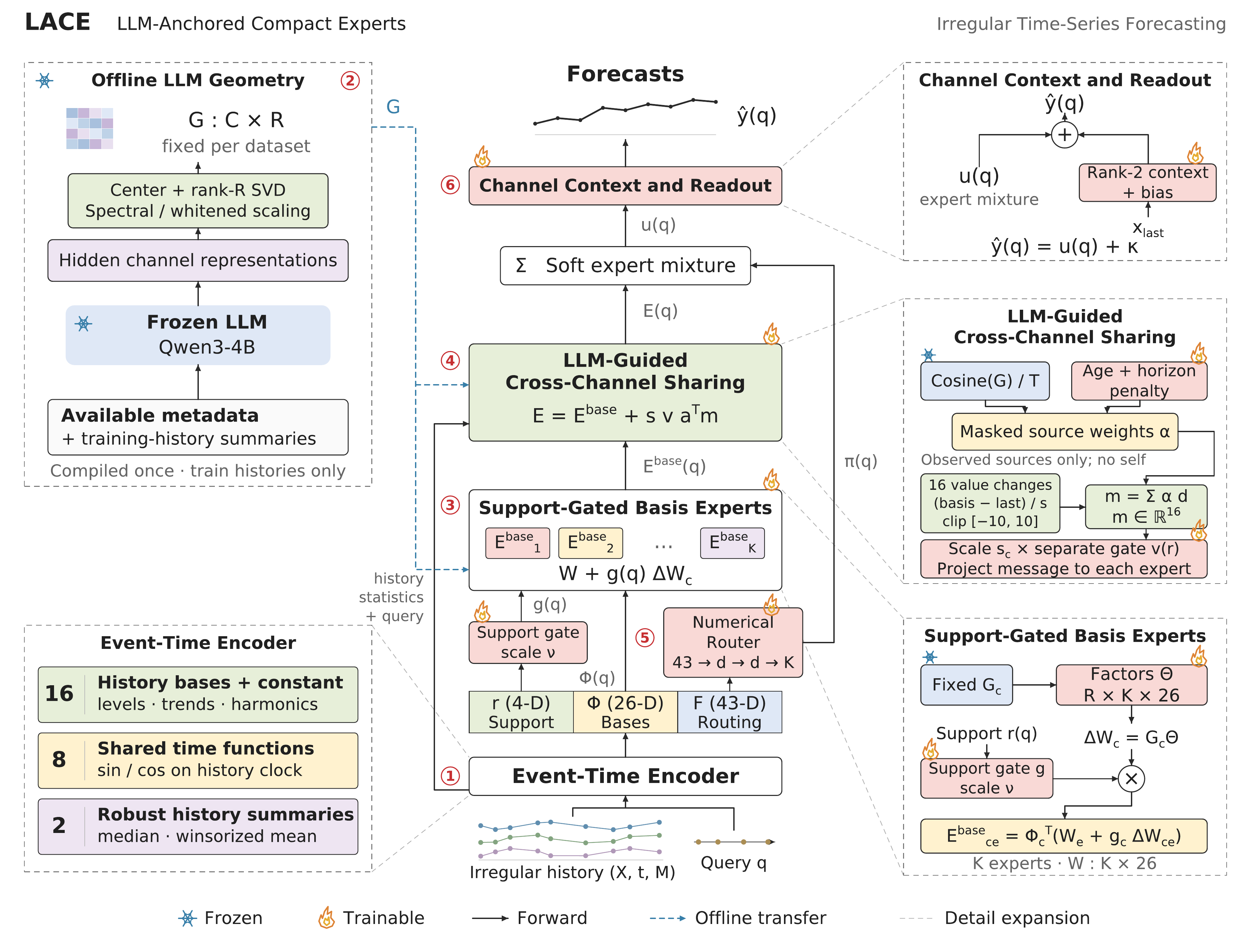}
  \caption{\textbf{LACE with LLM-guided cross-channel sharing.}
  (1) The encoder maps observed events and query times to temporal bases, routing features, and support statistics.
  (2) A frozen LLM computes channel coordinates $G\in\mathbb{R}^{C\times R}$ offline.
  These coordinates guide (3) support-gated adjustments to the expert coefficients and (4) sharing of standardized changes from other observed channels.
  Each recipient discounts stale source observations, restores the shared message to its own scale, and gates its contribution. Equation~\eqref{eq:lace-standardized-changes} defines the 16-dimensional changes.
  (5) The numerical router mixes the resulting expert forecasts.
  (6) A rank-two context adds a correction based on the last observed values.
  $C$, $K$, $R$, and $d$ denote the channel count, expert count, geometry rank, and router width.
  Blue dashed arrows carry fixed geometry. Gray dashed lines connect modules to their detail panels. Snowflakes mark frozen components and flames mark trainable components.}
  \label{fig:lace-framework}
\end{figure*}
% The six module headings match the optimized framework figure.
\section{Method}
\label{sec:lace-method}

Figure~\ref{fig:lace-framework} shows how LACE predicts a value at a requested time. An event-time encoder builds temporal bases from each channel's observations. A frozen LLM encodes the channels once, before training. The resulting coordinates determine channel-specific adjustments to the shared expert coefficients and supply the similarities used to exchange forecast changes across channels. Gates and a router use the observed history to set the strength of each contribution at prediction time. 

\subsection{Problem Formulation}
\label{sec:lace-problem}

Let $X\in\mathbb{R}^{L\times C}$ denote a history at timestamps $t_1,\ldots,t_L$, with observation mask $M\in\{0,1\}^{L\times C}$. Each channel may be observed at a different subset of these times. Given the history start $t_s$, cutoff $t_0$, and query time $q\geq t_0$, LACE predicts the channel values at $q$ from the observed history and the query timestamp. At training and evaluation time, the target mask identifies the entries to score.

With $D=\max(t_0-t_s,10^{-5})$, define
\begin{equation}
 \tau_i=\frac{t_i-t_s}{D},\qquad h(q)=\frac{q-t_0}{D},\qquad
 \delta_c(q)=a_c+h(q),
 \label{eq:lace-clock}
\end{equation}
where $a_c=1-\tau_{\mathrm{last},c}$ is the age of the latest observation in channel $c$. These quantities distinguish the forecast horizon from the time since that channel was last measured.

\subsection{Event-Time Encoder}
\label{sec:lace-time}

The encoder summarizes the observed events at their original timestamps. Midpoints between consecutive observations partition the history interval. Each event receives the duration of its interval as a weight. Four query-dependent windows give local means and damped extrapolations. Together with full-history summaries, harmonic predictions, time functions, and robust statistics, they form the basis vector $\boldsymbol\Phi_c(q)\in\mathbb{R}^{26}$.

The numerical router receives 35 history descriptors, four features for the query and observation support, and four standardized local deviations, for 43 features in total. Both gates receive $\mathbf{r}_c(q)=[1,\log(1+n_c)/5,a_c,h(q)]$, where $n_c$ is the channel's observation count. The encoder has no learned parameters. Appendix~\ref{app:lace-encoder} gives the equations and constants.

\subsection{Offline LLM Geometry}
\label{sec:lace-geometry}

We use the frozen Qwen3-4B-Instruct-2507 LLM~\cite{qwen2025instruct} to encode each channel. The prompt pairs available domain metadata with a profile of its training histories. Ten statistics describe level, within-history variation, variation in level across histories, count, roughness of consecutive changes, linear fit, four-cycle harmonic fit, repeated-value concentration, observation gap, and missingness. Channels whose names are unknown have anonymous identifiers. The prompt asks for a short description of a temporal pattern the channel might share with others and a reason it might differ.

We encode the prompt and its generated description together and average the final-layer hidden states of the last 16 tokens. Let $A\in\mathbb{R}^{C\times H}$ collect these channel vectors after subtracting their mean across channels. We take the first $R$ directions of $A=U\Sigma V^\top$ and scale the coordinates as follows.
\begin{equation}
 G=\begin{cases}
  \sqrt{C}\,U_R, & \text{whitened},\\[2pt]
  \displaystyle\frac{\sqrt{CR}\,U_R\Sigma_R}{\|\Sigma_R\|_F}, & \text{spectral}.
 \end{cases}
 \label{eq:lace-geometry}
\end{equation}
Both scalings give unit root-mean-square entries. Whitening weights all retained directions equally. Spectral scaling preserves the relative singular values. Within each dataset, validation selects the rank among the available nonzero directions under a fixed scaling convention.

Row $\mathbf{g}_c$ of $G\in\mathbb{R}^{C\times R}$ represents channel $c$ across histories and query times. The coefficient and cross-channel paths both read this row. Once compiled, $G$ is stored with the forecaster. The same channel descriptions can be re-encoded if a higher rank is needed.

\subsection{Support-Gated Basis Experts}
\label{sec:lace-experts}

Each of the $K$ experts combines all 26 bases. A trainable tensor $\Theta\in\mathbb{R}^{R\times K\times26}$ maps the channel coordinates to offsets from the shared coefficients $W\in\mathbb{R}^{K\times26}$.
\begin{align}
 \Delta W_{cej}&=\sum_{r=1}^{R}G_{cr}\Theta_{rej},\\
 g_c(q)&=\frac{\operatorname{sigmoid}(\mathbf{w}_g^\top\mathbf{r}_c(q)+b_g)}{\sqrt{1+n_c/\nu}},\\
 E^{\mathrm{base}}_{ce}(q)&=\sum_{j=1}^{26}\Phi_{cj}(q)
  [W_{ej}+g_c(q)\Delta W_{cej}].
 \label{eq:lace-experts}
\end{align}
The ungated coefficient offsets lie in the $R$-dimensional channel subspace spanned by $G$. The factor $(1+n_c/\nu)^{-1/2}$ bounds the gate from above. This bound decreases with the observation count $n_c$ at a rate controlled by the support scale $\nu>0$. The learned sigmoid also uses observation age and forecast horizon. A single gate scales all coefficient offsets for each channel and query.

We initialize $\Theta$ to zero and each shared expert head to a local mean or an average of local means. The average starting forecast across experts is the same for each candidate $K$. Training then learns how far to move the coefficients from their shared values for each channel.

\subsection{LLM-Guided Cross-Channel Sharing}
\label{sec:lace-sharing}

%The second path lets observations of one channel inform a forecast for another. We exchange predicted changes so that each recipient can use them on its own scale. 
Each source channel contributes estimates of how far its future value will rise or fall from its last observation. LACE divides each predicted change by the source channel’s historical variability so that changes can be compared across channels. It then combines the changes from the other channels and scales the forecast adjustment by the target channel’s own variability.
The index set $\mathcal{J}$ selects 16 measurement-valued bases from $\boldsymbol\Phi$, comprising two full-history means, four local means, four extrapolations, four fitted harmonic predictions, and two robust summaries.

For source channel $j$, define
\begin{align}
 s_j&=\max\!\left\{\sqrt{\operatorname{Var}_{v}(x_j)+10^{-6}},\,0.1\right\},\\
 \mathbf{d}_j(q)&=\operatorname{clip}_{[-10,10]}
  \left(\frac{\boldsymbol\Phi_{j,\mathcal{J}}(q)-x_{\mathrm{last},j}\mathbf{1}}{s_j}\right).
 \label{eq:lace-standardized-changes}
\end{align}
Here $\operatorname{Var}_{v}$ weights observations by their normalized interval lengths. Centering each basis prediction at the last observed value and dividing by $s_j$ expresses the change in units of the source channel's historical variability.

For a recipient $c$, the available sources are $\mathcal{N}_c=\{j:j\neq c,\ n_j>0\}$. We normalize geometry rows using a norm floor of $10^{-8}$ and compute
\begin{align}
 e_{cj}(q)&=\frac{\langle\bar{\mathbf{g}}_c,\bar{\mathbf{g}}_j\rangle}{T}
  -\operatorname{softplus}(\gamma)\frac{a_j+h(q)}{\bar\Delta_j+0.01},\\
 \alpha_{cj}(q)&=\frac{\exp e_{cj}(q)}{\sum_{k\in\mathcal{N}_c}\exp e_{ck}(q)},\quad j\in\mathcal{N}_c,\\
 \mathbf{m}_c(q)&=\sum_{j\in\mathcal{N}_c}\alpha_{cj}(q)\mathbf{d}_j(q).
 \label{eq:lace-messages}
\end{align}
For an empty source set we take $\mathbf{m}_c(q)=\mathbf{0}$. The temperature $T>0$ scales the contribution of channel similarity. The learned nonnegative penalty reduces the weight of a source as the elapsed time since its last observation grows relative to its typical gap.

An expert-specific projection and a separate gate map this message onto the recipient's scale.
\begin{align}
 v_c(q)&=\operatorname{sigmoid}(\mathbf{w}_v^\top\mathbf{r}_c(q)+b_v),\\
 E_{ce}(q)&=E^{\mathrm{base}}_{ce}(q)+s_c v_c(q)\mathbf{a}_e^\top\mathbf{m}_c(q),
 \label{eq:lace-shared-experts}
\end{align}
Here $\mathbf{a}_e\in\mathbb{R}^{16}$. We initialize $\mathbf{a}_e$ and both gate affine parameters in Eq.~\eqref{eq:lace-shared-experts} to zero, giving an initial cross-channel correction of zero. The source-age penalty starts at $\operatorname{softplus}(\gamma)=0.1$.

\subsection{Numerical Router}
\label{sec:lace-router}

A shared MLP with dimensions $43\rightarrow d\rightarrow d\rightarrow K$ and GELU activations after the first two layers computes the mixture weights.
\begin{equation}
 \boldsymbol\pi_c(q)=\operatorname{softmax}\bigl(\operatorname{MLP}(\mathbf{F}_c(q))+\mathbf{b}^{\pi}_c\bigr).
 \label{eq:lace-mixture}
\end{equation}
The numerical features $\mathbf F_c(q)$ and channel-specific bias $\mathbf b^\pi_c$ determine the weight of each expert after coefficient adaptation and cross-channel sharing. Initializing the final projection and routing biases to zero gives uniform initial weights.

\subsection{Channel Context and Readout}
\label{sec:lace-readout}

A rank-two numerical context adds a level adjustment shared across query times within a history.
\begin{align}
 \boldsymbol\kappa&=U_2D_2\mathbf{x}_{\mathrm{last}}+\mathbf{b}^{\mathrm{out}},\\
 \widehat y_c(q)&=\sum_{e=1}^{K}\pi_{ce}(q)E_{ce}(q)+\kappa_c.
 \label{eq:lace-output}
\end{align}
The bias-free matrices are $D_2\in\mathbb{R}^{2\times C}$ and $U_2\in\mathbb{R}^{C\times2}$. We initialize $U_2$ to zero and combine the mixture and the context in the units of the preprocessed data.

\subsection{Training and Implementation}
\label{sec:lace-implementation}

We train with the observed-target MSE in each nonempty minibatch,
\begin{equation}
 \mathcal{L}_{\mathcal B}=\frac{\sum_{b,q,c}M^y_{bqc}(\widehat y_{bqc}-y_{bqc})^2}{\sum_{b,q,c}M^y_{bqc}}.
 \label{eq:lace-loss}
\end{equation}
AdamW updates the expert coefficients, gates, message projection, source-age penalty, router, and rank-two context. We train for up to 200 epochs with early-stopping patience ten. Reported test metrics pool observed targets within each seed and average over seeds 2024--2028. Appendix~\ref{app:lace-protocol} gives the dataset-specific settings.

For the adopted multi-expert models, the number of trainable forecasting parameters is
\begin{equation}
 \begin{aligned}
 P_{\mathrm{train}}={}&d^2+45d+K(d+C+43+26R)\\
 &+5C+11.
 \end{aligned}
 \label{eq:lace-parameters}
\end{equation}
The offline teacher has 4,022,468,096 parameters. Each forecaster also stores $CR+4$ fixed scalars for geometry and frequencies. The implementation keeps a frozen $R\times d\times K$ routing tensor outside the forward computation. Table~\ref{tab:lace-structure} gives trainable and stored parameter counts for each dataset.

For a batch of $B$ histories with $L$ timestamps and $Q$ queries, computing the query features costs $O(BCLQJ)$ for $J=4$ windows. Given cached features, routing and basis projection cost $O(BCQ(43d+d^2+dK+26K))$, and the geometry-conditioned coefficients cost $O(26CRK)$. The dense graph requires $O(C^2R)$ operations for channel similarities, $O(16BC^2Q)$ for message aggregation, and $O(BC^2Q)$ memory for attention weights.
\section{Experiments}
\label{sec:lace-experiments}
\begin{table*}[!t]
\centering
\caption{Forecasting error on four irregular time-series datasets (mean $\pm$ standard deviation). LACE results use the sample standard deviation over five seeds (2024--2028). Bold and underline mark the lowest and second-lowest means. $\dagger$ denotes repaired PhysioNet.}
\label{tab:accuracy}
\setlength{\tabcolsep}{3pt}
\renewcommand{\arraystretch}{1.25}
\resizebox{\textwidth}{!}{%
\begin{tabular}{@{}lcccccccc@{}}
\toprule
Method & \multicolumn{2}{c}{HumanActivity} & \multicolumn{2}{c}{USHCN} & \multicolumn{2}{c}{PhysioNet $^\dagger$} & \multicolumn{2}{c}{MIMIC-III} \\
\cmidrule(lr){2-3}\cmidrule(lr){4-5}\cmidrule(lr){6-7}\cmidrule(lr){8-9}
 & MSE $\downarrow$ & MAE $\downarrow$ & MSE $\downarrow$ & MAE $\downarrow$ & MSE $\downarrow$ & MAE $\downarrow$ & MSE $\downarrow$ & MAE $\downarrow$ \\
\midrule
PrimeNet & $4.2507\pm0.0041$ & $1.7018\pm0.0011$ & $0.4930\pm0.0015$ & $0.4954\pm0.0018$ & $0.8128\pm0.0000$ & $0.6977\pm0.0001$ & $0.9073\pm0.0001$ & $0.6614\pm0.0001$ \\
NeuralFlows & $0.1722\pm0.0090$ & $0.3150\pm0.0094$ & $0.2087\pm0.0258$ & $0.3157\pm0.0187$ & $0.3910\pm0.0030$ & $0.4360\pm0.0020$ & $0.6085\pm0.0101$ & $0.5306\pm0.0066$ \\
CRU & $0.1387\pm0.0073$ & $0.2607\pm0.0092$ & $0.2168\pm0.0162$ & $0.3180\pm0.0248$ & $0.5917\pm0.0034$ & $0.5604\pm0.0020$ & $0.5895\pm0.0092$ & $0.5151\pm0.0048$ \\
mTAN & $0.0993\pm0.0026$ & $0.2219\pm0.0047$ & $0.5561\pm0.2020$ & $0.5015\pm0.0968$ & $0.3711\pm0.0046$ & $0.4231\pm0.0036$ & $0.9408\pm0.1126$ & $0.6755\pm0.0459$ \\
SeFT & $1.3786\pm0.0024$ & $0.9762\pm0.0007$ & $0.3345\pm0.0022$ & $0.4083\pm0.0084$ & $0.7912\pm0.0022$ & $0.6860\pm0.0008$ & $0.9230\pm0.0015$ & $0.6628\pm0.0008$ \\
GNeuralFlow & $0.3936\pm0.1585$ & $0.4541\pm0.0841$ & $0.2205\pm0.0421$ & $0.3286\pm0.0412$ & $0.8433\pm0.1546$ & $0.6791\pm0.0652$ & $0.8957\pm0.0209$ & $0.6450\pm0.0072$ \\
GRU-D & $0.1893\pm0.0627$ & $0.3253\pm0.0485$ & $0.2097\pm0.0493$ & $0.3045\pm0.0305$ & $0.3270\pm0.0016$ & $0.3917\pm0.0009$ & $0.4759\pm0.0100$ & $0.4526\pm0.0055$ \\
Raindrop & $0.0916\pm0.0072$ & $0.2114\pm0.0072$ & $0.2035\pm0.0336$ & $0.3029\pm0.0264$ & $0.3330\pm0.0016$ & $0.3965\pm0.0014$ & $0.6754\pm0.1829$ & $0.5444\pm0.0868$ \\
Warpformer & $0.0449\pm0.0010$ & $0.1228\pm0.0018$ & $0.1888\pm0.0598$ & $0.2939\pm0.0591$ & $0.2888\pm0.0014$ & $0.3556\pm0.0010$ & $0.4302\pm0.0035$ & $0.4025\pm0.0014$ \\
tPatchGNN & $0.0443\pm0.0009$ & $0.1247\pm0.0031$ & $0.1885\pm0.0403$ & $0.3084\pm0.0479$ & $0.2923\pm0.0029$ & $0.3595\pm0.0032$ & $0.4431\pm0.0115$ & $0.4077\pm0.0088$ \\
GraFITi & $0.0437\pm0.0005$ & $0.1221\pm0.0017$ & $0.1691\pm0.0093$ & $0.2777\pm0.0248$ & $\underline{0.2873}\pm0.0021$ & $\underline{0.3531}\pm0.0021$ & $0.4359\pm0.0455$ & $0.4142\pm0.0297$ \\
APN & $0.0421\pm0.0001$ & $0.1159\pm0.0006$ & $\underline{0.1590}\pm0.0137$ & $\mathbf{0.2611}\pm0.0167$ & $0.2947\pm0.0031$ & $0.3610\pm0.0061$ & $\underline{0.4292}\pm0.0027$ & $\underline{0.4016}\pm0.0016$ \\
KAFNet & $\underline{0.0418}\pm0.0002$ & $\underline{0.1140}\pm0.0007$ & $0.1916\pm0.0191$ & $0.2909\pm0.0338$ & $0.2963\pm0.0017$ & $0.3630\pm0.0043$ & $0.4472\pm0.0302$ & $0.4052\pm0.0211$ \\
\midrule
LACE & $\mathbf{0.0404}\pm0.0002$ & $\mathbf{0.1131}\pm0.0004$ & $\mathbf{0.1548}\pm0.0027$ & $\underline{0.2632}\pm0.0019$ & $\mathbf{0.2862}\pm0.0008$ & $\mathbf{0.3513}\pm0.0015$ & $\mathbf{0.4063}\pm0.0021$ & $\mathbf{0.3847}\pm0.0004$ \\
\bottomrule
\end{tabular}%
}
\par\vspace{2pt}\begin{minipage}{\textwidth}\footnotesize Ranks use the available precision before rounding to four decimal places.\end{minipage}
\end{table*}

We evaluate LACE on four forecasting benchmarks, then use ablations to measure the contributions of its LLM-conditioned coefficients and cross-channel sharing. 

\subsection{Experimental Setup}
\label{sec:lace-experimental-setup}

\paragraph{Datasets and evaluation.}
HumanActivity contains localization measurements~\citep{vidulin2010activity}, and USHCN contains climate records from the forecasting benchmark of \citet{debrouwer2019gruode}. The two clinical benchmarks use the PhysioNet/Computing in Cardiology Challenge 2012 data~\cite{silva2012physionet} and MIMIC-III~\cite{johnson2016mimic}. We follow APN's preprocessing and data partitions for HumanActivity, USHCN, and MIMIC-III. We use the repaired hourly PhysioNet data with the original cohort and partitions, computing normalization statistics on the training partition. Appendix~\ref{app:lace-protocol} gives the sample counts and forecast windows.

For each test query, LACE receives its timestamp and the observed history. We compute MSE and MAE by pooling squared and absolute errors, respectively, over observed forecast targets. For LACE and its variants, we report the mean and sample standard deviation across five training seeds (2024--2028) on the same partition.

\paragraph{Baselines.}
Table~\ref{tab:accuracy} compares LACE with PrimeNet~\cite{chowdhury2023primenet}, NeuralFlows~\cite{bilos2021neuralflows}, CRU~\cite{schirmer2022cru}, mTAN~\cite{shukla2021mtan}, SeFT~\cite{horn2020seft}, GNeuralFlow~\cite{mercatali2024gneuralflow}, GRU-D~\cite{che2018grud}, Raindrop~\cite{zhang2022raindrop}, Warpformer~\cite{zhang2023warpformer}, tPatchGNN~\cite{zhang2024tpatchgnn}, GraFITi~\cite{yalavarthi2024grafiti}, APN~\citep{liu2026apn}, and KAFNet~\citep{zhou2026kafnet}. Relative improvements use the baseline means and the unrounded LACE means.

\paragraph{Training and selection.}
We train LACE and its ablations with AdamW~\citep{loshchilov2019adamw} for up to 200 epochs, with early-stopping patience of ten epochs. Each ablation uses the same criterion as the full model on that dataset.

All four datasets use the same architecture, with 26 temporal bases, 43 routing features, and a channel geometry computed offline by the frozen Qwen3-4B-Instruct-2507 LLM. We tune the expert count $K$, router width $d$, support scale $\nu$, geometry rank $R$, and graph temperature $T$ for each dataset. Within each dataset, we fix the batch size, the source of the hidden representations, and the convention for scaling the geometry. We compare configurations by five-seed mean validation MSE and stop when changing any one hyperparameter on its grid gives no further improvement. We fixed the final model and ablation settings before evaluating their test errors. Appendix~\ref{app:lace-protocol} details the search grids and computational cost.

\subsection{Main Results}
\label{sec:lace-main-results}
\begin{table*}[!t]
\centering
\caption{Ablations and channel-assignment controls. We report test means $\pm$ sample standard deviations for the same five seeds. Each variant uses the full model's hyperparameters for its dataset. $\dagger$ denotes repaired PhysioNet.}
\label{tab:lace-ablations}
\setlength{\tabcolsep}{3pt}
\renewcommand{\arraystretch}{1.25}
\resizebox{\textwidth}{!}{%
\begin{tabular}{@{}lcccccccc@{}}
\toprule
Variant & \multicolumn{2}{c}{HumanActivity} & \multicolumn{2}{c}{USHCN} & \multicolumn{2}{c}{PhysioNet $^\dagger$} & \multicolumn{2}{c}{MIMIC-III} \\
\cmidrule(lr){2-3}\cmidrule(lr){4-5}\cmidrule(lr){6-7}\cmidrule(lr){8-9}
 & MSE $\downarrow$ & MAE $\downarrow$ & MSE $\downarrow$ & MAE $\downarrow$ & MSE $\downarrow$ & MAE $\downarrow$ & MSE $\downarrow$ & MAE $\downarrow$ \\
\midrule
Full LACE & $0.0404\pm0.0002$ & $0.1131\pm0.0004$ & $0.1548\pm0.0027$ & $0.2632\pm0.0019$ & $0.2862\pm0.0008$ & $0.3513\pm0.0015$ & $0.4063\pm0.0021$ & $0.3847\pm0.0004$ \\
w/o LLM branch & $0.0405\pm0.0003$ & $0.1132\pm0.0005$ & $0.1678\pm0.0043$ & $0.2799\pm0.0027$ & $0.2857\pm0.0007$ & $0.3507\pm0.0016$ & $0.4071\pm0.0025$ & $0.3863\pm0.0015$ \\
w/o Cross-Channel Sharing & $0.0406\pm0.0003$ & $0.1133\pm0.0004$ & $0.1629\pm0.0033$ & $0.2702\pm0.0018$ & $0.2864\pm0.0008$ & $0.3517\pm0.0011$ & $0.4063\pm0.0027$ & $0.3855\pm0.0017$ \\
Shuffled LLM geometry & $0.0408\pm0.0004$ & $0.1134\pm0.0005$ & $0.1606\pm0.0031$ & $0.2695\pm0.0037$ & $0.2860\pm0.0004$ & $0.3512\pm0.0010$ & $0.4069\pm0.0016$ & $0.3870\pm0.0019$ \\
Non-semantic graph, w/o LLM & $0.0405\pm0.0004$ & $0.1128\pm0.0007$ & $0.1603\pm0.0036$ & $0.2748\pm0.0036$ & $0.2856\pm0.0008$ & $0.3506\pm0.0009$ & $0.4053\pm0.0026$ & $0.3850\pm0.0008$ \\
\bottomrule
\end{tabular}%
}

\end{table*}
\begin{figure*}[!t]
\centering
\includegraphics[width=\textwidth]{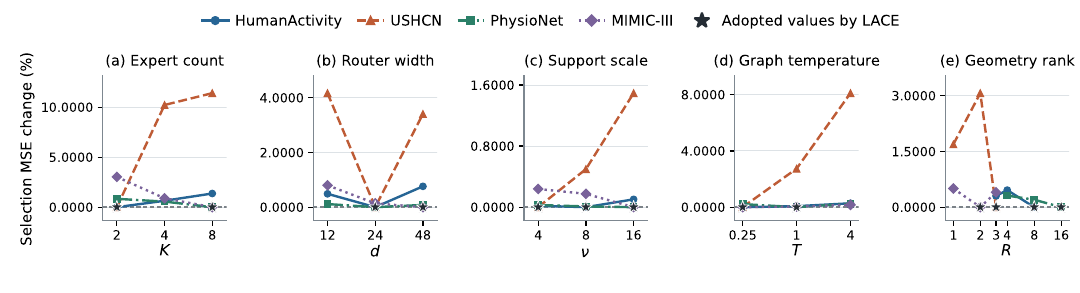}
\caption{Validation sensitivity to one hyperparameter at a time. Curves show the percentage change in five-seed mean validation MSE relative to each dataset's adopted configuration. Stars mark the adopted settings. Datasets with the same setting share a star.}
\label{fig:lace-sensitivity}
\end{figure*}

LACE has the lowest mean MSE on all four datasets and the lowest mean MAE on HumanActivity, PhysioNet, and MIMIC-III (Table~\ref{tab:accuracy}). Relative to APN, MSE falls by 4.0055\%, 2.6392\%, 2.8701\%, and 5.3293\% on HumanActivity, USHCN, PhysioNet, and MIMIC-III, respectively. MIMIC-III shows the largest reduction, from 0.4292 to 0.4063. MAE on this dataset also falls from 0.4016 to 0.3847, a 4.2036\% improvement.

On HumanActivity, LACE improves on the closest competitor, KAFNet, by 3.3166\% in MSE and 0.7675\% in MAE. On PhysioNet, the gap to GraFITi is narrower. LACE achieves an MSE of 0.2862 and an MAE of 0.3513, compared with 0.2873 and 0.3531 for GraFITi. On USHCN, LACE achieves an MSE of $0.1548\pm0.0027$, compared with 0.1590 for APN. APN leads on MAE with 0.2611, followed by LACE at $0.2632\pm0.0019$, a gap of 0.8168\%.

\subsection{Module Ablations}
\label{sec:lace-module-ablations}

We retrain each variant with the full model's dataset-specific hyperparameters, optimizer, checkpoint rule, and five seeds (Table~\ref{tab:lace-ablations}). Removing the LLM branch disables both the geometry-conditioned coefficient offsets and the exchange of forecast changes between channels. Removing cross-channel sharing alone retains the coefficient offsets.

The LLM branch contributes most on USHCN. Removing it raises mean MSE by 8.4056\%, and the full model has lower MSE in all five seeds. Removal also raises mean MSE on HumanActivity and MIMIC-III. PhysioNet shows a small gain of 0.1927\% after removal. Cross-channel sharing has a more consistent effect on absolute errors. Removing it raises mean MAE on all four datasets.

On USHCN, assigning the same geometry vectors to shuffled channels raises mean MSE by 3.7346\% at the same model capacity. The full model also outperforms a graph whose edge weights depend only on historical reliability in all five seeds. Appendix~\ref{app:lace-ablation-analysis} reports the per-dataset comparisons.

\subsection{Validation Sensitivity}
\label{sec:lace-sensitivity}

Figure~\ref{fig:lace-sensitivity} varies one of $K,d,\nu,T,R$ at a time around the adopted configuration for each dataset. The selected value of each hyperparameter attains or ties the lowest five-seed mean validation MSE along its sweep. We evaluate MAE at the same checkpoints. Appendix~\ref{app:lace-sensitivity} gives absolute validation errors and standard deviations.

The selected models use two experts on HumanActivity and USHCN and eight on PhysioNet and MIMIC-III. Across the values tested, changing the expert count yields a wider range of validation MSE than changing the support scale on every dataset.

\subsection{Parameter Budget and Scope}
\label{sec:lace-experimental-scope}

LACE uses 2,301 trainable forecasting parameters on HumanActivity, 1,992 on USHCN, 5,999 on PhysioNet, and 6,867 on MIMIC-III. The 4,022,468,096-parameter teacher compiles each dataset's channel representations offline. Section~\ref{sec:lace-implementation} gives the cost of the forecasting operations. Table~\ref{tab:lace-structure} lists stored parameter counts.

Figure~\ref{fig:lace-efficiency} compares the accuracy and computational cost of five forecasters. Across the four datasets, LACE has the fewest trainable parameters on average (4.29K). Its matrix products and convolutions require 48.24M FLOPs per batch, compared with 1.11G for APN. Extra inference memory averages 24.44\,MiB for LACE and 115.22\,MiB for APN. APN has the shortest training and inference times, at 13.13 and 1.51\,ms per batch, respectively. LACE takes 17.34 and 5.94\,ms. KAFNet has the lowest matrix and convolution FLOPs and the smallest extra inference allocation when its released HumanActivity architecture is used on all four datasets. Appendix~\ref{app:lace-efficiency} gives the workload, configurations, and costs for each dataset.

Cross-channel sharing scales quadratically with the number of channels, which ranges from 5 to 96 in these experiments. For channels with verified names, the teacher can use those names alongside the numerical histories. For the remaining channels, it uses anonymous indices and numerical profiles.\space\space

\section{Conclusion}
\label{sec:lace-conclusion}

LACE uses a frozen LLM to compile one representation per channel. The forecaster then uses these representations to adapt expert coefficients and share forecast changes across channels. Both operations depend on the support available in the observed history. Once the representations are stored, forecasting uses only the compact numerical model.

Across four benchmarks, the same architecture uses 1,992--6,867 trainable forecasting parameters and achieves the lowest mean MSE among the methods evaluated. LACE also achieves the lowest mean MAE on three datasets and ranks second on USHCN. Retrained ablations place the largest contribution of the LLM branch on USHCN, where LACE outperforms every control on both metrics in all five seeds. The branch also lowers mean MSE on HumanActivity and MIMIC-III. On PhysioNet, removing it yields a slightly lower mean MSE. 
\label{end:main-body}

\section*{Impact Statement}

LACE reduces the trainable parameter count of an irregular time-series forecaster, which may make forecasting easier to deploy in settings with limited computing resources. The clinical experiments measure prediction
error on retrospective benchmarks. Use in patient care would require prospective validation, including checks across patient groups, sites, and changes in measurement practice. Channel representations can reflect patterns
in how measurements were collected, and errors in sparsely observed channels may disproportionately affect groups with less complete records.

\bibliography{example_paper}
\bibliographystyle{icml2027}

\clearpage
\appendix
\onecolumn
\section{Dataset and Optimization Details}
\label{app:lace-protocol}

% Required packages: booktabs, graphicx. Appendix is one-column.
\begin{table}[!ht]
\centering
\caption{Forecasting instances by partition. The last column counts observed test targets across all channels. $\dagger$ denotes repaired PhysioNet.}
\label{tab:lace-data}
\small
\setlength{\tabcolsep}{9pt}
\renewcommand{\arraystretch}{1.15}
\begin{tabular}{@{}lrrrrr@{}}
\toprule
Dataset & Channels & Train & Validation & Test & Test targets \\
\midrule
HumanActivity & 12 & 949 & 193 & 218 & 5,910 \\
USHCN & 5 & 901 & 101 & 112 & 368 \\
PhysioNet $^\dagger$ & 36 & 9,703 & 1,079 & 1,199 & 27,008 \\
MIMIC-III & 96 & 17,212 & 1,913 & 2,125 & 23,015 \\
\bottomrule
\end{tabular}
\end{table}

\begin{table}[!ht]
\centering\small
\caption{Optimizer settings for Full and its ablations. Learning rates are constant. The last column identifies the learning-rate multiplier for the coefficient tensor $\Theta$ or the shared expert coefficients $W$. All other parameters use the base rate.}
\label{tab:lace-optimization}
\begin{tabular}{@{}llrrrrl@{}}
\toprule
Dataset & Optimizer & Base LR & Weight decay & Batch & Clip & LR multiplier \\
\midrule
HumanActivity & AdamW & 0.001 & 0.001 & 16 & -- & $\Theta$: 0.25 \\
USHCN & AdamW & 0.0003 & 0.001 & 32 & 0.1 & All: 1 \\
PhysioNet & AdamW & 0.001 & 0.01 & 32 & 0.1 & All: 1 \\
MIMIC-III & AdamW & 0.001 & 0.001 & 32 & -- & $W$: 0.25 \\
\bottomrule
\end{tabular}
\end{table}
\begin{table}[!ht]
\centering\small
\caption{Dataset configurations selected on validation MSE. $K,d,R,\nu,T$ denote expert count, router width, geometry rank, support scale, and graph temperature. The two parameter columns give the trainable and total stored forecaster weights.}
\label{tab:lace-structure}
\begin{tabular}{@{}lrrrrrlrr@{}}
\toprule
Dataset & $K$ & $d$ & $R$ & $\nu$ & $T$ & Geometry scaling & Trainable & Stored \\
\midrule
HumanActivity & 2 & 24 & 8 & 8 & 0.25 & Spectral & 2,301 & 2,685 \\
USHCN & 2 & 24 & 3 & 4 & 0.25 & Whitened & 1,992 & 2,136 \\
PhysioNet & 8 & 24 & 16 & 16 & 1 & Spectral & 5,999 & 9,071 \\
MIMIC-III & 8 & 48 & 2 & 16 & 0.25 & Whitened & 6,867 & 7,635 \\
\bottomrule
\end{tabular}
\end{table}

\paragraph{Benchmark partitions.}
Table~\ref{tab:lace-data} gives the number of forecasting instances in each partition and the number of observed targets in each test set. We use APN's preprocessing and partitions for HumanActivity, USHCN, and MIMIC-III. HumanActivity splits records before constructing windows and uses the stored position values. USHCN follows the benchmark's ID split, and MIMIC-III uses sequential ID partitions. Both use APN's preprocessed measurements. All seeds share the same partitions.

\paragraph{Forecast queries.}
HumanActivity uses 3,000\,ms of history and a 300\,ms forecast window, with a 4,000\,ms stride. USHCN predicts the first three retained timestamps after 149.5 on a 200-unit clock. PhysioNet uses observations in $[0,36)$ hours and predicts the earliest three retained hourly bins in $[36,48)$ hours. MIMIC-III uses 30-minute bins and predicts the next three retained timestamps after 71.5 on its 96-bin clock.

\paragraph{PhysioNet repair and missing values.}
The repaired PhysioNet cache uses the original cohort and partitions. Preprocessing resolves the Troponin I/T field aliases, marks negative raw measurements as missing, and aggregates observations by hour. We standardize all partitions with the training means and population standard deviations and retain observations with absolute standardized values below five. The training, validation, and test sets contain 217,124, 24,061, and 27,008 observed forecast targets, respectively. Across all four datasets, observation masks determine which values enter encoder statistics and losses.

\paragraph{Optimization and checkpoint selection.}
Table~\ref{tab:lace-optimization} lists the settings shared by Full and its ablations on each dataset. AdamW uses $\beta=(0.9,0.999)$, $\epsilon=10^{-8}$, and a constant learning rate. We apply weight decay to tensors with at least two dimensions. The shared expert head and geometry coefficient tensor use the learning-rate multipliers in the table. All other parameter groups use the base rate.

We train for up to 200 epochs with early-stopping patience ten. On HumanActivity, PhysioNet, and MIMIC-III, the checkpoint score averages masked MSE over validation batches. USHCN pools squared error over observed validation targets. Ties reset patience and select the later checkpoint. Across the five Full fits per dataset, the selected epochs range from 30--44 on HumanActivity, 42--58 on USHCN, 13--66 on PhysioNet, and 26--38 on MIMIC-III. The longest runs lasted 54, 68, 76, and 48 epochs, respectively. Test MSE and MAE pool errors over observed targets on each dataset.

USHCN has 334 observed validation targets, of which eighteen fall in the last minibatch. Averaging the four minibatch scores gives that batch one quarter of the weight. Pooling gives each target equal weight. We trained each candidate in the final USHCN search with pooled MSE for both checkpoint and configuration selection.

\paragraph{Architecture and configuration selection.}
We first searched over $K,d,\nu,T,R$ on all four datasets, keeping each dataset's batch size, optimizer settings, and geometry scaling fixed. We ranked configurations by mean validation batch-MSE over seeds 2024--2028. The search started from 44 previously evaluated dataset--configuration pairs (220 seed runs). Crossing the current settings with values that improved MSE in single-parameter sweeps gave 27 candidate combinations and required sixty additional fits.

At each subsequent step we evaluated the center and ten neighbors obtained by varying one coordinate at a time. We updated the center when a neighbor reduced the unrounded five-seed mean. For ties among improving candidates we used trainable parameter count, then configuration identifier. A search ended when the center matched or improved on every neighbor. The initial combinations and the three subsequent rounds required 60, 140, 70, and 35 new fits, respectively. The 305 new fits and 220 existing runs covered 105 dataset--configuration pairs. Eighty additional ablation and control fits brought the first stage to 385 new training runs, followed by one hundred test evaluations. We use this stage's results for HumanActivity, PhysioNet, and MIMIC-III.

For USHCN we repeated selection with pooled MSE and included the learning rate and weight decay in the search. The initial 36 configurations combined $d\in\{12,24\}$, $R\in\{1,2,3\}$, learning rate in $\{0.0003,0.001\}$, and weight decay in $\{0.001,0.01,0.1\}$, with $K=2$, $\nu=4$, and $T=0.25$. We then varied each structural or optimization parameter in turn using the same update and tie rules. Across this stage, 43 configurations required 215 training runs, five per configuration. Twenty ablation and control fits brought the total to 235 runs. We evaluated the five selected Full checkpoints and the twenty control checkpoints on the test set. Tables~\ref{tab:lace-optimization} and~\ref{tab:lace-structure} give the selected settings.

\paragraph{Final evaluation.}
Configuration selection and final evaluation use seeds 2024--2028. We fixed the configurations, code, channel geometry, and checkpoints before scoring each stage's final models. The reported results comprise 75 model--seed evaluations from the first stage and 25 from the USHCN search. Earlier experiments had also used these test partitions. Sample standard deviations (denominator four) describe training variability across seeds on the fixed splits.

\paragraph{Offline geometry and parameter accounting.}
All four datasets use the final-layer mean over the last 16 tokens from frozen Qwen3-4B-Instruct-2507, given the training-history prompt and its generated channel description. Channels with unverified names use anonymous identifiers. HumanActivity and PhysioNet use spectral scaling and unit entry RMS. USHCN and MIMIC-III use whitened coordinates. We select the rank from each dataset's grid.

Table~\ref{tab:lace-structure} reports both trainable and stored parameter counts. Stored counts include the implementation's frozen $R\times d\times K$ routing tensor. We count the $CR+4$ fixed buffer values and the 4,022,468,096 parameters of the offline teacher separately.

\paragraph{Sensitivity grids.}
The structural grid uses $K\in\{2,4,8\}$, $d\in\{12,24,48\}$, $\nu\in\{4,8,16\}$, and $T\in\{0.25,1,4\}$. The candidate ranks are $\{3,4,8\}$ for HumanActivity, $\{4,8,16\}$ for PhysioNet, and $\{1,2,3\}$ for USHCN and MIMIC-III. For each rank we take the leading directions of the same frozen representation. We truncate the whitened coordinates directly and normalize the spectral coordinates to unit entry RMS after truncation.

Each of the four datasets contributes five structural sweeps around its selected configuration. The resulting 60 plotted points share the center across the five sweeps on each dataset, giving 44 distinct dataset--configuration pairs and 220 seed runs.

\paragraph{Reproducibility.}
Training uses Python 3.12, PyTorch 2.8.0 with CUDA 12.8, and NumPy 2.3.2. LLM extraction uses Transformers 4.51.3 and BF16. The forecaster uses FP32 with the TF32 setting recorded in the run manifest.

\begin{table}[!ht]
\centering\small
\caption{Trainable forecaster parameters by ablation. The corresponding numbers of stored parameter scalars, including disabled tensors, are 2,685, 2,136, 9,071, and 7,635 in dataset order.}
\label{tab:lace-ablation-capacity}
\setlength{\tabcolsep}{7pt}
\begin{tabular}{@{}lrrrr@{}}
\toprule
Variant & HumanActivity & USHCN & PhysioNet & MIMIC-III \\
\midrule
Full LACE & 2,301 & 1,992 & 5,999 & 6,867 \\
w/o LLM branch & 1,842 & 1,793 & 2,532 & 6,312 \\
w/o Cross-Channel Sharing & 2,263 & 1,954 & 5,865 & 6,733 \\
Shuffled LLM geometry & 2,301 & 1,992 & 5,999 & 6,867 \\
Non-semantic graph, w/o LLM & 1,880 & 1,831 & 2,666 & 6,446 \\
\bottomrule
\end{tabular}
\end{table}

\subsection{Additional Ablation Analysis}
\label{app:lace-ablation-analysis}

\paragraph{LLM-conditioned paths.}
The \emph{w/o LLM branch} variant removes the geometry-conditioned coefficient offsets, cross-channel sharing, and their gates. It retains the shared experts, numerical router, channel context, and output biases. Every ablation uses the Full model's hyperparameters and is trained from scratch. Parameter counts appear in Table~\ref{tab:lace-ablation-capacity}. On USHCN, branch removal raises MSE by 8.4056\% and MAE by 6.3369\%, with Full achieving lower errors in all five seeds. Mean MSE rises by 0.1582\% on HumanActivity and 0.1850\% on MIMIC-III. MIMIC-III MAE rises by 0.3996\%, with the same direction in every seed. On PhysioNet, branch removal lowers mean MSE by 0.1927\% and improves MSE in all five seeds.

\paragraph{Cross-channel sharing.}
Removing cross-channel sharing retains the geometry-conditioned coefficients and their support gate. Mean MAE increases on every dataset. USHCN has the largest increase (2.6398\%) and a 5.2258\% rise in mean MSE, with both errors higher in every seed. Mean MSE also rises on HumanActivity (0.4553\%) and PhysioNet (0.0381\%), with Full ahead in four of five seeds on each. On MIMIC-III, removing sharing lowers mean MSE by 0.0031\%. Both means round to 0.4063.

\paragraph{Channel assignment and non-semantic sharing.}
\emph{Shuffled LLM geometry} applies a fixed channel permutation to the geometry used by both paths, preserving parameter count. For each dataset we generate the permutation with seed 2024 and use it across all five training seeds. On USHCN, shuffling raises mean MSE by 3.7346\% and MAE by 2.3656\%, with Full ahead on both metrics in every seed. Its mean MSE also rises by 0.9420\% on HumanActivity and 0.1415\% on MIMIC-III, and falls by 0.0878\% on PhysioNet.

The \emph{non-semantic graph, w/o LLM} control removes coefficient offsets and their support gate and assigns a constant semantic similarity to every source. Historical reliability then determines the graph weights, and the model still learns the message projection and the graph gate. On USHCN this control raises mean MSE by 3.5693\% and MAE by 4.3827\%. Full has lower errors in every seed. Full also has lower mean MSE on HumanActivity. The control has lower mean MSE on both clinical datasets and lower mean MAE on HumanActivity.
\clearpage
\section{Additional Sensitivity Results}
\label{app:lace-sensitivity}

\begin{figure}[!ht]
\centering
\includegraphics[width=.98\textwidth,height=.67\textheight,keepaspectratio]{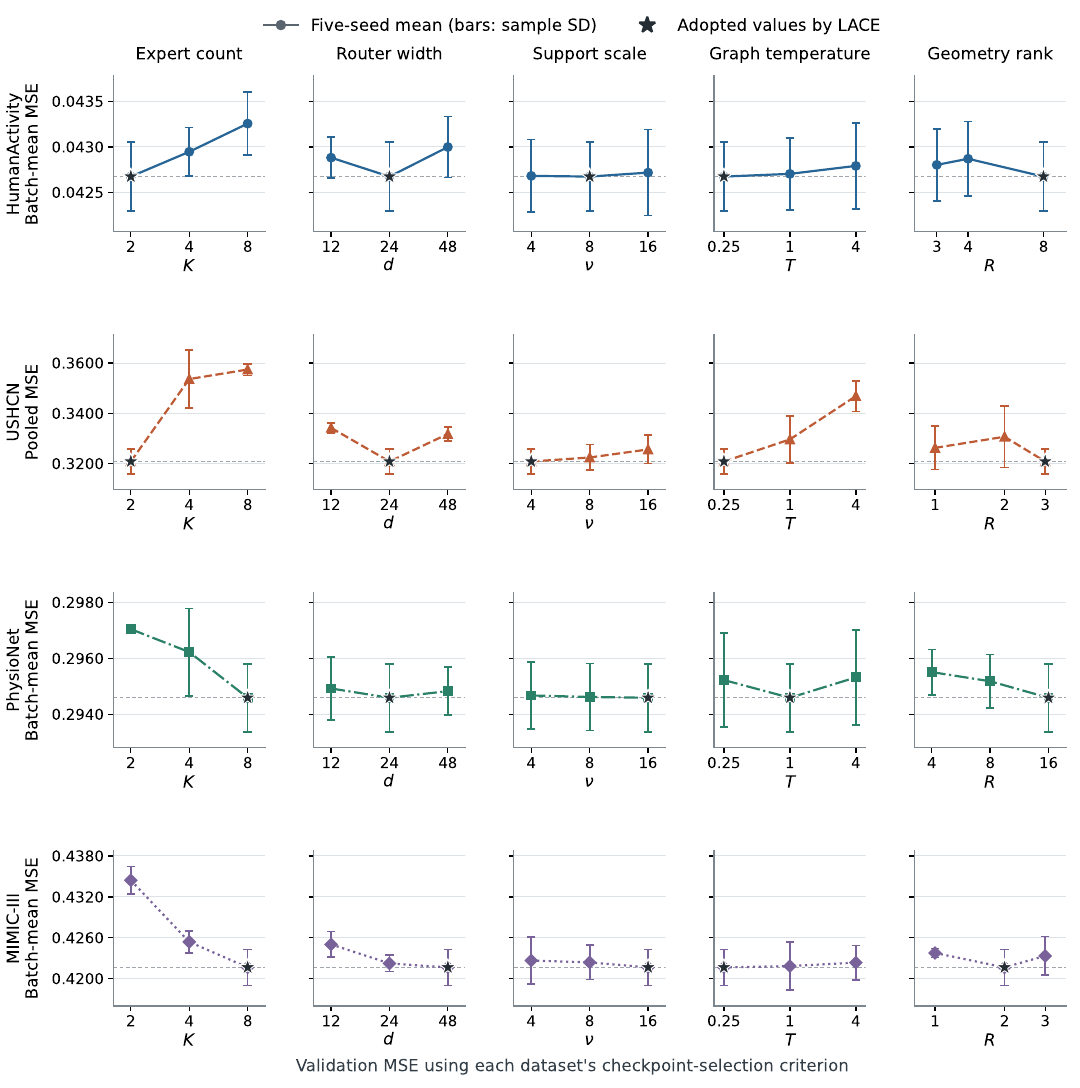}
\caption{Validation MSE as each parameter varies around the selected setting. Stars mark the selected values and the horizontal dashed line shows their mean MSE. Bars show one sample standard deviation across five seeds. Panels within a row share a vertical scale.}
\label{fig:lace-sensitivity-mse-full}
\end{figure}

\begin{figure}[!ht]
\centering
\includegraphics[width=.98\textwidth,height=.73\textheight,keepaspectratio]{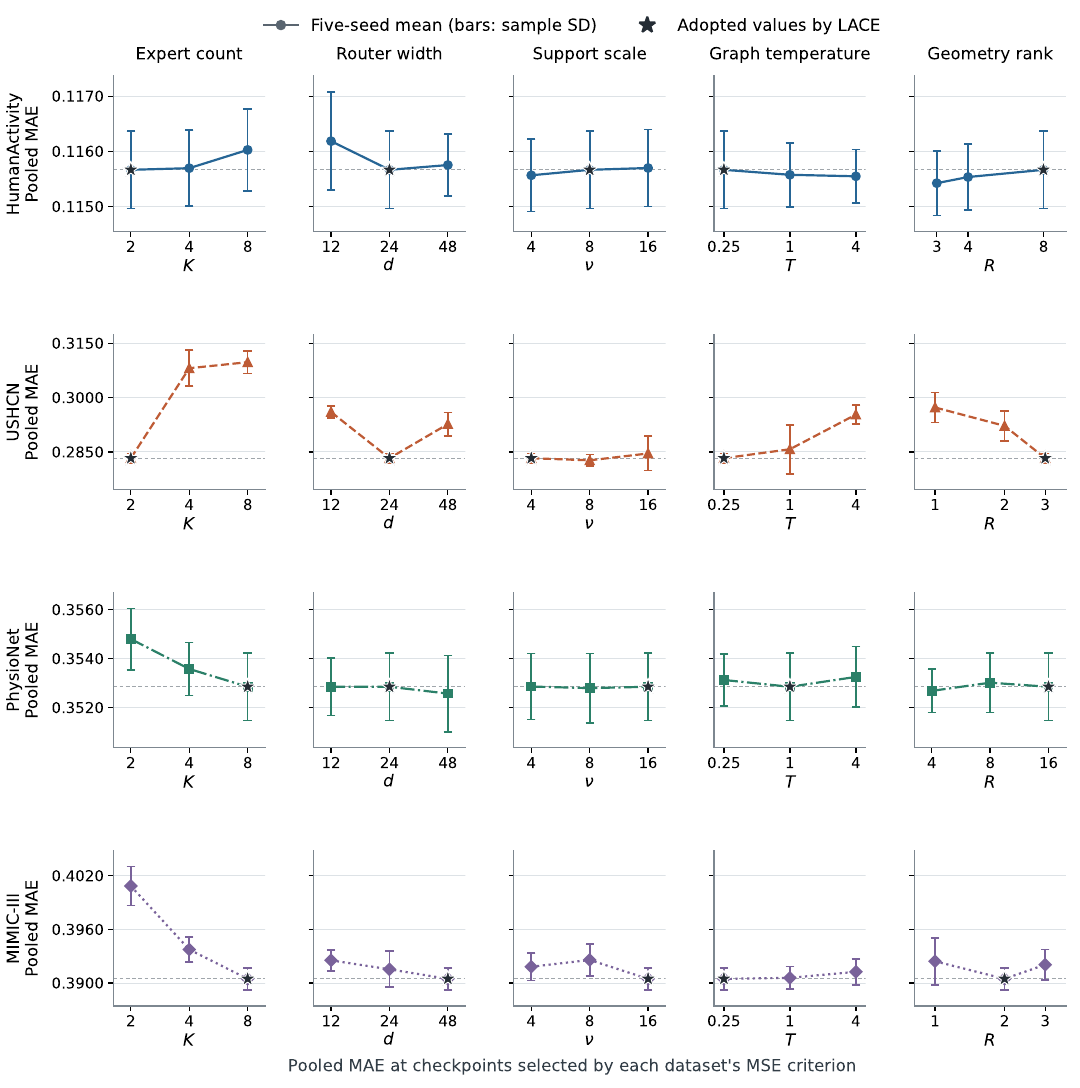}
\caption{Validation MAE at the checkpoints used for Figure~\ref{fig:lace-sensitivity-mse-full}. Stars mark the MSE-selected values. Bars show one sample standard deviation across five seeds. Panels within a row share a vertical scale.}
\label{fig:lace-sensitivity-mae}
\end{figure}

Figures~\ref{fig:lace-sensitivity-mse-full} and~\ref{fig:lace-sensitivity-mae} show how validation error changes when a single parameter varies around the selected configuration. We report the mean and sample standard deviation over seeds 2024--2028. For MSE, USHCN pools squared errors over observed targets and the other datasets average masked MSE over fixed batches. For MAE, each dataset pools absolute errors over observed targets at the MSE-selected checkpoints.

Stars identify the settings selected by five-seed mean validation MSE. At these settings, the mean selection MSE is 0.0427 on HumanActivity, 0.3209 on USHCN, 0.2946 on PhysioNet, and 0.4217 on MIMIC-III.

\paragraph{Expert capacity and routing width.}
The final neighborhoods favor two experts on HumanActivity and USHCN, and eight on the clinical datasets. Increasing $K$ from two to eight raises selection MSE by 1.3653\% on HumanActivity and 11.3973\% on USHCN. Reducing $K$ from eight to two raises it by 0.8336\% on PhysioNet and 3.0282\% on MIMIC-III. The selected router width is 24 on HumanActivity, USHCN, and PhysioNet and 48 on MIMIC-III. Doubling USHCN's width from 24 to 48 increases error by 3.4113\%, whereas reducing MIMIC-III's width from 48 to 12 increases it by 0.7999\%.

\paragraph{Geometry and support.}
The selected ranks are 8, 3, 16, and 2 for HumanActivity, USHCN, PhysioNet, and MIMIC-III, respectively. On MIMIC-III, ranks one and three raise validation batch-MSE by 0.5017\% and 0.4001\% relative to rank two. Graph temperature has its largest effect on USHCN, where increasing $T$ from 0.25 to 4 raises pooled validation MSE by 8.1093\%. Varying the support scale raises selection MSE by at most 0.1021\%, 1.4935\%, 0.0279\%, and 0.2383\% on the four datasets, respectively.

On USHCN, increasing the support scale from $\nu=4$ to $\nu=8$ lowers validation MAE from 0.2833 to 0.2827 and raises pooled validation MSE from 0.3209 to 0.3225.
\clearpage
\section{Event-Time Encoder Details}
\label{app:lace-encoder}

Midpoints between consecutive observations of a channel divide its normalized history interval $[0,1]$ into segments. Each observed event receives a temporal weight $v_{ic}$ equal to its segment length. Masked entries receive zero weight. A mean computed with these weights describes the value over elapsed time. A counting mean gives equal weight to each measurement.

At each of six exponential rates $\{0,1,4,16,64,256\}$, we compute the temporal mean, clipped slope, standard deviation, difference between counting and temporal means, and log effective count. We append the last value, its age, the mean observed gap $\bar\Delta_c$, log count, and an observation-validity flag to form the 35-dimensional history descriptor $\mathbf{s}_c$. We clip slopes to $[-20,20]$ and divide them by 20, and divide log counts by five.

The query also defines four local windows. For $\alpha_j\in\{1/4,1,4,16\}$, we use
\begin{align}
 \lambda_{cjq}&=\frac{\alpha_j}{\delta_c(q)+\bar\Delta_c+0.01},\\
 w_{icjq}&\propto v_{ic}\exp[-\lambda_{cjq}(\tau_{\mathrm{last},c}-\tau_i)].
 \label{eq:lace-windows}
\end{align}
Weights are normalized over the observed history. Let $\mu_{cjq}$ and $\bar\tau_{cjq}$ be the corresponding value and time means. Each window yields a mean and a damped local extrapolation,
\begin{align}
 s_{cjq}&=\frac{\operatorname{Cov}_w(\tau,x)}{\operatorname{Var}_w(\tau)+10^{-3}+0.1\bar\Delta_c^2},\\
 \ell_{cjq}&=\mu_{cjq}+\frac{s_{cjq}z_{cjq}}{1+\lambda_{cjq}z_{cjq}},\quad
 z_{cjq}=1+h(q)-\bar\tau_{cjq}.
 \label{eq:lace-local}
\end{align}
Four additional predictions come from separate harmonic regressions with regressors $[1,\sin(2\pi f\tau),\cos(2\pi f\tau)]$, $f\in\{1,2,3,4\}$. Each fit uses count-normalized observed sums and ridge $0.02I$, and is evaluated at $1+h(q)$.

The 26 entries of $\boldsymbol\Phi_c(q)$ comprise the last observation, full-history temporal and counting means, four local means, four extrapolations, four fitted harmonic predictions, a constant, eight sine and cosine terms, and two robust summaries. The robust summaries are the observed median and the mean after clipping observations to the median plus or minus three median absolute deviations. The harmonics and trigonometric functions use normalized time within the history.

The numerical routing vector is
\begin{align}
 \mathbf{z}_c(q)&=\left[h,\delta_c,\frac{\log(1+n_c)}5,
  \frac{\min\{\delta_c/(\bar\Delta_c+0.01),20\}}{10}\right],\\
 \boldsymbol\eta_c(q)&=\operatorname{clip}_{[-10,10]}
  \left(\frac{\boldsymbol\mu_c(q)-x_{\mathrm{last},c}}{\sigma_c^{\mathrm{count}}}\right),\\
 \mathbf{F}_c(q)&=[\mathbf{s}_c;\mathbf{z}_c(q);\boldsymbol\eta_c(q)]\in\mathbb{R}^{43},
 \label{eq:lace-routing-input}
\end{align}
where $(\sigma_c^{\mathrm{count}})^2$ is the observed counting variance plus $0.01$. Both gates use the four-dimensional support vector
\begin{equation}
 \mathbf{r}_c(q)=[1,\log(1+n_c)/5,a_c,h(q)].
 \label{eq:lace-support}
\end{equation}
We compute these fixed features outside the gradient calculation. A channel with no observations has zero value summaries, unit age and mean gap, and a zero validity flag. Its counting variance includes the same regularizer used for observed channels. The time functions and constant basis are defined for every channel.
\clearpage
\section{Efficiency Benchmark}
\label{app:lace-efficiency}

\paragraph{Workload.}
We sample 96 distinct records from each test partition by a fixed permutation and group them into three batches of 32. All models use the same batches, including the repaired PhysioNet data. Histories and queries are padded within each batch. Timing starts with these tensors on the GPU and includes each model's input preparation, such as LACE's event encoding, tPatchGNN's patch construction, and GraFITi's graph construction from the requested query coordinates. The timed training steps use target values to compute masked MSE. We initialize separate model instances for profiling.

\paragraph{Configurations.}
We use the selected forecasting configurations for LACE and the authors' implementations for the four baselines. APN uses the official dataset-specific launch settings. GraFITi uses the settings published for the benchmark protocol of \citet{zhang2024tpatchgnn}. For tPatchGNN we divide the shared history window into the official number of equal-duration patches for each dataset. KAFNet's repository provides a configuration for HumanActivity. We use this configuration on all four datasets, adjusting the channel count to match each input.

\paragraph{Measurement.}
We use a single NVIDIA RTX 4090 D with PyTorch 2.8.0, CUDA 12.8, and FP32 with TF32 enabled. For inference and training separately, we run three warm-up passes over each batch followed by twelve timed passes using the default kernels. We synchronize the GPU before and after each timed step and report mean wall-clock time per batch. Each training step includes the forward pass, masked loss, backpropagation, and an AdamW update. To measure extra inference memory, we subtract the allocation for resident models and inputs from the peak allocation for each batch and average the three differences.

We count matrix products and 2D convolutions with the PyTorch operator profiler and 1D convolutions from their output shapes, counting two FLOPs per multiply-add. Unfused attention exposes its matrix products to the profiler. The timing measurements also include FFTs, linear solves, sorting, scatter, and elementwise operations. Parameter counts include all forecaster scalars with gradients enabled. LACE loads its precomputed channel geometry before profiling.

\begin{table}[!ht]
\centering
\caption{Efficiency by dataset on an RTX 4090 D. FLOPs count matrix products and convolutions in a forward pass. Times and memory are per batch of 32.}
\label{tab:lace-efficiency-per-dataset}
\small
\begin{tabular}{@{}llrrrrr@{}}
\toprule
Dataset & Model & Trainable params. & FLOPs ($10^6$) & Train (ms) & Infer. (ms) & Extra (MiB) \\
\midrule
HumanActivity & LACE & 2,301 & 19.99 & 16.14 & 6.02 & 31.31 \\
 & APN & 28,789 & 3292.12 & 13.87 & 2.18 & 221.28 \\
 & GraFITi & 249,409 & 5165.03 & 42.36 & 5.29 & 52.73 \\
 & tPatchGNN & 165,634 & 1399.98 & 26.79 & 2.78 & 89.54 \\
 & KAFNet & 5,135 & 21.91 & 24.07 & 1.60 & 4.94 \\
\midrule
USHCN & LACE & 1,992 & 6.07 & 17.40 & 6.23 & 11.40 \\
 & APN & 1,970 & 306.45 & 11.37 & 1.17 & 62.71 \\
 & GraFITi & 495,617 & 11139.51 & 28.21 & 3.56 & 130.34 \\
 & tPatchGNN & 167,542 & 1132.89 & 23.19 & 2.96 & 167.37 \\
 & KAFNet & 5,135 & 16.20 & 22.26 & 1.81 & 5.90 \\
\midrule
PhysioNet & LACE & 5,999 & 24.60 & 17.99 & 5.72 & 11.48 \\
 & APN & 6,701 & 144.93 & 14.75 & 1.23 & 22.58 \\
 & GraFITi & 250,945 & 5931.20 & 41.77 & 5.15 & 48.73 \\
 & tPatchGNN & 362,178 & 6111.38 & 23.96 & 3.38 & 563.25 \\
 & KAFNet & 5,135 & 32.44 & 25.60 & 2.08 & 5.54 \\
\midrule
MIMIC-III & LACE & 6,867 & 142.29 & 17.84 & 5.78 & 43.58 \\
 & APN & 13,777 & 713.27 & 12.55 & 1.45 & 154.30 \\
 & GraFITi & 254,785 & 5545.77 & 40.13 & 5.56 & 56.00 \\
 & tPatchGNN & 363,378 & 27068.90 & 30.02 & 11.37 & 2969.90 \\
 & KAFNet & 5,135 & 103.01 & 27.58 & 2.21 & 22.05 \\
\bottomrule
\end{tabular}
\end{table}
\end{document}